# Seed Classification using Synthetic Image Datasets Generated from Low-Altitude UAV Imagery

Venkat Margapuri
Department of Computer Science
Kansas State University
Manhattan, KS
marven@ksu.edu

Niketa Penumajji
Department of Computer Science
Kansas State University
Manhattan, KS
niketa912@ksu.edu

Mitchell Neilsen
Department of Computer Science
Kansas State University
Manhattan, KS
neilsen@ksu.edu

*Abstract*— Plant breeding programs extensively monitor the evolution of seed kernels for seed certification, wherein lies the need to appropriately label the seed kernels by type and quality. However, the breeding environments are large where the monitoring of seed kernels can be challenging due to the minuscule size of seed kernels. The use of unmanned aerial vehicles aids in seed monitoring and labeling since they can capture images at low altitudes whilst being able to access even the remotest areas in the environment. A key bottleneck in the labeling of seeds using UAV imagery is drone altitude i.e. the classification accuracy decreases as the altitude increases due to lower image detail. Convolutional neural networks are a great tool for multi-class image classification when there is a training dataset that closely represents the different scenarios that the network might encounter during evaluation. However, with the seeds being in a breeder environment coupled with the varying image resolution and clarity, it is challenging to generate a training dataset that covers every evaluation scenario. The article addresses the challenge of training data creation using Domain Randomization wherein synthetic image datasets are generated from a meager sample of seeds captured by the bottom camera of an autonomously driven Parrot AR Drone 2.0. Besides, the article proposes a seed classification framework as a proof-of-concept using the convolutional neural networks of Microsoft's ResNet-100, Oxford's VGG-16, and VGG-19. To enhance the classification accuracy of the framework, an ensemble model is developed resulting in an overall accuracy of 94.6%. The task of classification is performed on five different types of seeds, canola, rough rice, sorghum, soy, and wheat.

## I. INTRODUCTION

Seed Phenotyping is the comprehensive assessment of complex seed traits such as growth, development, tolerance, resistance, ecology, yield, and the measurement of parameters that form more complex traits [9] wherein one of the focus areas is seed certification. Seed certification, generally performed in plant breeding environments, is an internationally recognized system to maintain and make available to the public, high quality seed and propagating materials of superior adapted crop varieties grown and distributed to ensure varietal identity and purity. Seed certification is based on the premise that proper identification of crop varieties is essential to everyone who handles seed [21]. Proper labeling of the seeds is one of the aspects of seed certification which is often times challenging due to the voluminous nature of the breeding environments. Recent developments in the areas of deep learning have led to the evolution of powerful convolutional neural networks (CNN) that aid in the classification (or labeling) of a certain entity. CNNs perform best when they are trained on datasets that closely resemble the test criteria. However, the creation of a training dataset for seed labeling is challenging since seeds evolve leading to a change in their morphology. As a result, a universal dataset that resembles the evolution of seeds in a breeding environment is not available. More importantly, such datasets are required to be catered to a specific breeding environment. The creation of such datasets requires a profusion of seed sample availability which isn't always possible. The need for a large seed sample is addressed by the application of a technique named Domain Randomization (DR) i.e. the idea of training models on simulated images that transfer to real images. The use of DR means that a large synthetic dataset that resembles real-world seeds in different sizes and orientations may be generated using a meager sample of real seeds. The capture of the sample of seeds for DR may be automated by the use of autonomous unmanned aerial vehicles (UAV) aka drones. Drones can be flown autonomously at low altitudes and make good candidates to capture images in controlled plant breeding environments.

## II. RELATED WORK

The work by Yoda et. al. [20] applies the technique of domain randomization to train Mask R-CNN, an instance segmentation neural network on a synthetic image dataset of 20 different cultivars belonging to barley, four cultivars of wheat, and one cultivar each of rice, oat and lettuce. The average precision (AP) values computed based on mask regions at the varying intersection over union (IoU) thresholds achieved comparable AP50 values of 96% and 95% for the synthetic and real-world datasets respectively.

Ward, Moghadam, and Hudson [22] conducted similar experiments to perform automated segmentation of individual leaves of a plant for high-throughput phenotyping. The proposed technique leveraged synthetic images of plants generated from real plant images. The Mask R-CNN architecture was trained with a combination of real and synthetic images of Arabidopsis plants. The proposed technique achieved a 90% leaf segmentation score on the A1 test set and outperformed the state-of-the-art approaches for the CVPPP Leaf Segmentation Challenge (LSC).

Gulzar, Hamid, Soomro, Alwan and Journaux [7] developed a seed classification system for 14 different types of seeds employing the concept of transfer learning on a pre-trained model of VGG-16. The results showed a classification accuracy of 99% on a test image dataset of 234 images.

Buters, Belton, and Cross in [4] conducted experiments to perform seed and seedling detection and classification using unmanned aerial vehicles. The technique of object-based image analysis (OBIA) with eRecognition software was used as part of the experiments. The results showed the feasibility of low-cost commercially available UAVs in monitoring ecological recovery.

Coulibaly, Kamsu-Foguem, Kamissako, and Traore [5] proposed a deep learning technique with transfer learning in millet crop images using VGG-16. The goal of the study was to identify mildew disease in pearl millet. Fine-tuning was performed on a pre-trained model of VGG-16 by adding two fully connected layers and training the model on the image dataset. An accuracy of 95%, precision of 90.50%, recall of 94.5% and f1score of 91.75%.

In the study conducted by Mukti and Biswas [14], transfer learning-based plant disease detection was performed using ResNet-50. The dataset included 38 different classes of plant leaf images where in 70295 training images and 17572 validation images were used. Fine-tuning was performed to enhance the performance of the ResNet-50 model that eventually yielded a training accuracy of 99.80%. 33 images belonging to the 38 classes of leaves were used for testing. The proposed model yielded an accuracy of 100% on the test dataset.

Muneer and Fati [15] proposed a deep learning technique to classify and recognize the desired herb from thousands of herbs using shape and texture features. The proposed system employed two classifiers, Support Vector Machine (SVM) and Deep Learning Neural Network (DLNN). The models were tested on a dataset containing 1000 herbs. The results showed that the SVM model achieved a recognition accuracy of 74.63% whereas the DLNN achieved 93% accuracy. Furthermore, the processing time was four seconds for SVM and five seconds for DLNN.

A deep-learning classification system for identifying weeds using high-resolution UAV imagery was proposed by Bah, Dericquebourg, Hafiane, and Canals in [2]. The proposed system was applied to images of vegetables captured about 20m above the soil using a UAV. The results showed that weed detection was effective in different crop fields with overall precisions of 93%, 81%, and 69% obtained for beet, spinach, and bean respectively.

Bristeau, Vissière, Callou and Petit [3] discuss the navigation and control technology embedded in the Parrot AR Drone. The article sheds light on the low-cost inertial sensors, computer vision techniques, attitude and velocity estimation, and control architecture used in the UAV.

III. SEED CLASSIFICATION FRAMEWORK

A. Parrot AR Drone 2.0

The Parrot AR Drone 2.0 is a quadcopter manufactured by the French company Parrot SA. The drone uses a 3630 OMAP CPU, a processor that is based upon a 32-bit ARM cortex and runs with 1 GHz. The drone uses 4 brushless motors running at 28.5 revolutions/minute which are controlled by an 8 MIPS AVR CPU on each motor controller. The drone has a front and a bottom camera to capture images and videos. The front camera provides an HD image resolution (720p-30fps) whereas the bottom camera provides QVGA (320 x 240) at 60fps. The drone may be controlled via a mobile application supported on Android and iOS provided by Parrot SA. Besides the mobile app, the quadcopter has support for autonomous navigation through libraries supported in Robot Operating System (ROS) and nodeJS. It turns out that javascript is a good fit to control drones autonomously for the work since it is inherently event-driven. The nodeJS library, node-ar-drone is leveraged to operate the drone autonomously in the breeding environment. While the node-ar-drone library provides a plethora of functions to operate the quadcopter, the ones used for the current work are take-off, land, spin, and video capture and stop. The behavior of the functions is self-intuitive from the names of the functions. The video captured by the drone is output in .H264 format when the video capture function of the node-ar-drone library is used. Videos in raw .H264 format don't carry rate information. However, different frames of the video are required to be extracted for the creation of synthetic images. As a work-around, the FFmpeg library is used to convert the video in .H264 format to .mp4 format rendering it conducive for frame extraction using the Python-OpenCV framework.

B. Domain Randomization Framework

A framework in Python-OpenCV to generate synthetic image datasets for DR is developed as part of the work and consists of the following steps:
1. Extract each of the frames from the video captured by the drone and identify a frame on which the seeds are visible clearly.
2. Load the frame into Gimp and extract the seeds on the image. Technically, the foreground on the image represents the seeds and may be extracted by separating the alpha channel from the image.
3. Pick a canvas onto which the extracted seeds may be laid. The canvas acts as the background for the image. The ideal canvas is the test site where the seeds are laid for phenotyping.
4. The seeds (extracted foregrounds) may now be laid on the canvas in different sizes and orientations that resemble the seeds in the phenotyping environment.

Image augmentation is applied to the generated synthetic seeds at random. Each of the seeds is put through a sequence of brightness adjustment, horizontal flip, vertical flip, rotation and scaling of varying degrees before being laid on the canvas.

## C. Synthetic Dataset

The seeds of canola, rough rice, sorghum, soy, and wheat are used as part of the experiment. 30 seeds belonging to each of the seed types in question i.e. canola, rough rice, sorghum, soy, and wheat are randomly selected from a large pool of available seeds belonging to each of the seed types and placed in the prototypical seed phenotyping environment. The Parrot AR Drone 2.0 is flown over the phenotyping environment autonomously and made to capture a video of the seeds as it moves across the seed phenotyping environment using its bottom camera. The drone is made to fly at three different heights of 0.3m, 0.5m, and 0.7m to capture three different videos. Four images of the white light-emitting lightbox are used as the canvas (background) for the synthetic images. While it is fair to assume that the lightbox looks the same across the board while emitting white light, it is not the case and minor discrepancies in the intensity are observed amongst lightboxes. As a result, using the images of different lightboxes ensures that minor discrepancies are captured during the creation of the synthetic dataset. Each of the videos is processed using the Python-OpenCV framework described in section b to create synthetic images for each of the seeds in question. The synthetic images generated are of the size 224 x 224 x3 to match the input size for the neural networks used to train on the dataset. For each of the seeds, 1000 synthetic images are generated of which the images consist of 20 - 50 seeds each. The dataset is made diverse by the fact that the images consist of seeds captured from different heights i.e. the 1000 images of a single seed comprise 333 images where the seeds are captured from a height of 0.3m, 333 images where the seeds are captured from a height of 0.5m and 334 images where the seeds are captured from a height of 0.7m. Overall, a total of 5000 images are generated amongst all of the seeds. One of the struggles in classifying multiple classes of seeds is the appearance of seeds at different heights. It is not an issue in cases where the images are captured from a fixed height. However, it is not always feasible whilst using drones since they are mobile and are influenced by environmental factors such as wind and obstacles. It is observed that a given seed appears smaller from larger heights and vice-versa. Owing to this, the training dataset consisting of all images from a fixed height may not be a representative sample of the real-world test set that the neural network might encounter. Hence, the drone is programmed to capture images from varying heights of 0.3m, 0.5m, and 0.7m to ensure that a real-world representative sample of seeds is obtained.

Fig. 1 depicts the workflow of the proposed seed classification framework, Fig. 2 shows the Parrot AR Drone 2.0 capturing images flying over the prototypical seed phenotyping environment and Fig. 3 shows the real and synthetic seeds used in the experiment.

## D. Neural Networks

Three state-of-the-art convolutional neural networks namely, Oxford's VGG-16, VGG-19, and Microsoft's ResNet-101 are leveraged as part of the work. The architectures of each of the networks are briefly described as follows.

*VGG-16:*

Broadly, the architecture comprises a stack of five convolutional layers followed by three fully connected layers. The input to the first convolutional layer is an RGB image of size 224 x 224 px. Each of the convolutional layers comprises a filter with a receptive field of 3 x 3. Max pooling layers are present after each of the convolutional layers to perform spatial pooling. Max pooling is performed over a 2 x 2-pixel window with a stride of 2. Of the three fully connected layers, the first two have a total of 4096 channels each, and the third dense layer has as many channels as the number of output categories that the network is expected to classify. The final layer is the softmax layer that outputs a probability of a certain image belonging to a certain class. ReLU is used as the activation function for each of the hidden layers.

*VGG-19:*

The architecture of VGG-19's is similar to that of VGG-16's. The architecture comprises a total of 19 layers in comparison to VGG-16's 16. Like VGG-16, the architecture of VGG-19 comprises a stack of five convolutional layers with max-pooling layers to reduce the spatial dimensions of the output volume. However, a total of 13 convolutional layers are observed in VGG-16 whereas 16 convolutional layers are present in VGG-19. The architecture follows VGG-16 in every other aspect.

*ResNet-101:*

The ResNet architecture relies on the idea of using a skip connection process that performs a convolution transformation

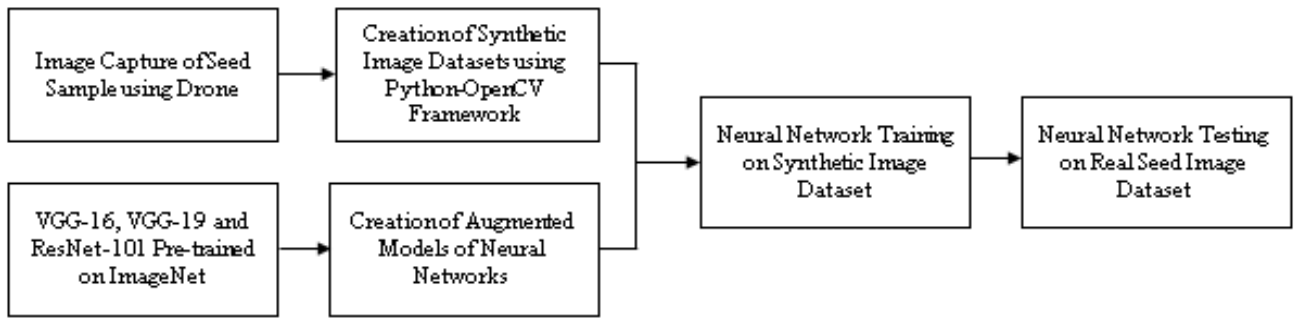

Fig. 1: Workflow of the Seed Classification System

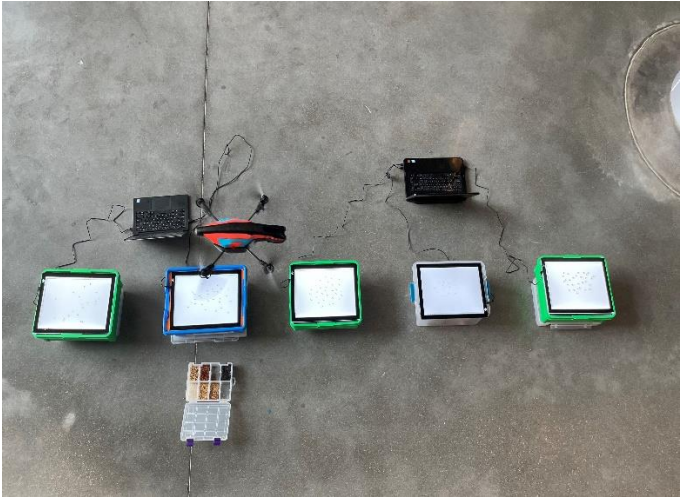

Fig. 2: Image Capture of Seeds in Phenotyping Environment using Parrot AR Drone 2.0

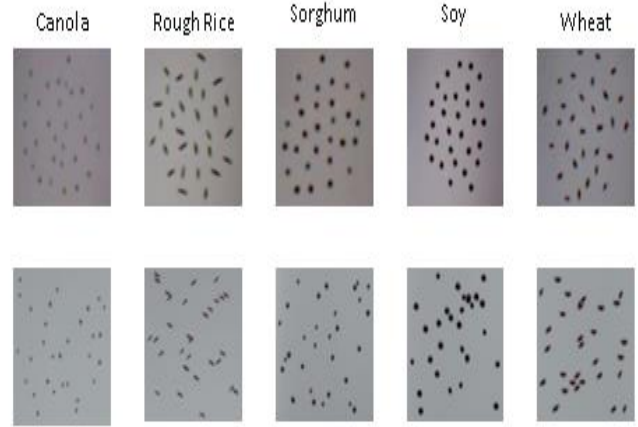

Fig. 3: Top row shows real seeds captured by the UAV; Bottom row shows synthetic images generated by the DR Framework using the real seeds

F(X) on an incoming feature X and adds the result to the original feature X. The modified feature is then served as the input to the next layer. It primarily solves the problem of vanishing gradient i.e. the condition where the loss function shrinks to zero after repeated applications of the chain rule. It is generally the case when the network architectures are overly deep. With ResNets, the network learns the residual elements. There are two main designs of residual elements, skip connection and identity mapping [11]. The training of the residual network may be explained as learning the residual function $Y = F(X) + X$ where the goal is to minimize $F(X)$ so it reaches 0. Applying L2 regularization or weight decay is a way to achieve $F(X) = 0$ since it incentivizes the network weights to be as small as possible. At that point, $Y = X$ i.e. an identity mapping. As a result, the addition of more layers to the network does not hurt the performance of the network. In other words, the presence of multiple residual blocks in the network allows the structure of the residual network to be self-regulated through skip connections thereby achieving the deepening effect of the network [11].

## IV. EXPERIMENT AND RESULTS

The three neural network architectures are imported from the canned architectures provided by Keras as part of their Applications module. The imported models are models pre-trained on the ImageNet dataset that comprises 1000 classes. The key idea behind the use of pre-trained models trained on ImageNet is that the notion of Transfer Learning may apply. Briefly, Transfer Learning is the ability of a neural network to apply the knowledge gained by training on a dataset to a different dataset where there is a presence of common domains between the source and target datasets. However, the ImageNet dataset does not consist of the five seeds or anything remotely similar that are of interest for the experiment. In cases where the source and target domains don't share domains, the sheer transfer of knowledge is usually unsuccessful and often results in poor performance. Since the neural network architectures are large, training the entire neural network from scratch requires an exorbitant amount of data. As a means to better train the network whilst preserving the learned features from the ImageNet dataset, the architecture of the neural networks is augmented. As for augmented architectures, three Sequential models (one for each of VGG-16, VGG-19, and ResNet-101) are built from the imported models wherein all of the layers except the fully connected layers and softmax layer are copied to the Sequential models. The VGG models consist of three fully connected layers whereas ResNet-101 has one. Three fully connected layers are added to the Sequential models followed by a softmax layer for classification. All of the transferred layers from the pre-trained models are rendered frozen for training and only the three fully connected layers added to the model are trained. The dataset used for the experiment is the synthetic image dataset described in III(c). 80% of the dataset is used for training and 20% of the dataset is used for validation. Experiments with different hyperparameters are conducted in the training phase. Upon training each model with a different set of hyperparameters, the models that yield the best accuracy and loss are saved. The hyperparameters used for each of the best models are as shown in Table 1.

Table 1: Hyperparameter Values for Training

| Hyperparameter | VGG-16 | VGG-19 | ResNet-101 |
| --- | --- | --- | --- |
| Nodes per trained layer | 512 | 512 | 1024 |
| Learning Rate | $1 \times e^{-3}$ | $1 \times e^{-3}$ | $1 \times e^{-2}$ |
| Learning Rate Decay | 100 steps @0.96 | 100 steps @0.96 | No Decay |
| Dropout | 0.4 | 0.5 | 0.5 |
| Batch Size | 32 | 32 | 32 |
| Optimizer | Adam | SGD | Adam |

The training accuracies and losses of the best models for the fine-tuned neural networks are as shown in Fig. 4. Overfitting is commonly observed across all three of the models. The use of Dropout helped with reducing the overfitting of the models. However, it results in the validation accuracy being higher than training accuracy in parts. Overall, the validation accuracy for each of the fine-tuned models of VGG-16, VGG-19, and ResNet-101 at the end of the training phase is 96.43%, 90.03%, and 97.81% respectively.

## V. EVALUATION

The image dataset for evaluation is generated from the images captured by the Parrot AR Drone 2.0 whilst flying over the seed phenotyping environment at different heights. While the images used during the training phase are generated from videos captured from pre-defined heights of 0.3m, 0.5m, and 0.7m, the images for the test dataset are generated from videos captured from varying heights of less than one meter. The image dataset is generated by extracting frames from videos using OpenCV. The images for the test dataset consist of 75 images containing 20 – 50 seeds of a certain seed type captured from varying heights. Overall, a test dataset consisting of 450 images is used to evaluate the performance of each of the networks. The results of evaluation based on the metrics of Accuracy, Precision, and Recall are as shown in Table 2. The metrics are briefly described below:

**Accuracy:** Accuracy is defined as the fraction of the samples in the dataset correctly classified by the classifier. It is given by $(true\ positives + true\ negatives)/(true\ positives + true\ negatives + false\ positives + false\ negatives)$.

**Precision:** Precision is given by $true\ positives/(true\ positives + false\ positives)$. Precision indicates the number of positives correctly classified by the classifier.

**Recall:** Recall is defined as the True Positive Rate of a classifier and given by $true\ positives/(true\ positives + false\ negatives)$. Recall indicates the number of positives detected by the classifier of all the positives in the dataset.

The results show that ResNet-101 lends itself best to the task of classification achieving an overall accuracy of 92% while VGG-16 and VGG-19 lag behind at 89% and 86% respectively. However, the performance of the all of the models is sub-par in comparison to the accuracy on the validation set during training.

## VI. ENSEMBLE MODEL

Ensemble Learning refers to the generation and combination of multiple inducers to solve a machine learning task [17]. In an attempt to better the classification performance, an ensemble whose predictions are based on the individual predictions made by each of the models is developed. The ensemble model works by summing the categorical softmax outputs of each of the models for a given sample. The prediction made by the model is the largest resultant categorical value. The workflow of the ensemble model is as shown in Fig. 5 and the results obtained are as shown in table 3. The results show an improvement in overall accuracy to 94.6% showing a 2.6% improvement in accuracy compared to the best individual model of ResNet-101. Besides, a significant improvement in the precision and recall for each of the individual classes is also observed which indicates better quality of predictions.

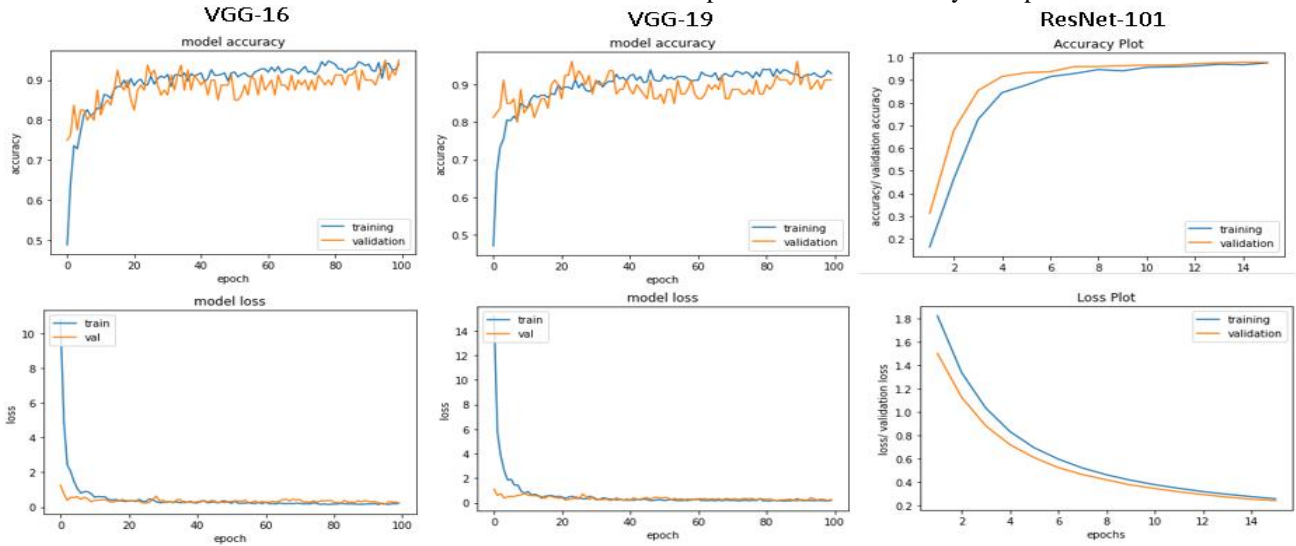

Fig. 4: Training and Validation Accuracies and Losses for Fine-Tuned VGG-16, VGG-19 and ResNet-101

Table 2: Evaluation Results on the Fine-Tuned Models of VGG-16, VGG-19 and ResNet-101

|  | VGG-16 | | | VGG-19 | | | ResNet-101 | | |
| --- | --- | --- | --- | --- | --- | --- | --- | --- | --- |
|  | Accuracy | Precision | Recall | Accuracy | Precision | Recall | Accuracy | Precision | Recall |
| Canola | 0.91 | 0.76 | 0.79 | 0.86 | 0.67 | 0.67 | 0.91 | 0.78 | 0.78 |
| Rough Rice | 0.91 | 0.78 | 0.78 | 0.90 | 0.77 | 0.80 | 0.93 | 0.81 | 0.87 |
| Sorghum | 0.90 | 0.70 | 0.77 | 0.75 | 0.66 | 0.60 | 0.91 | 0.78 | 0.77 |
| Soy | 0.90 | 0.75 | 0.80 | 0.90 | 0.74 | 0.77 | 0.93 | 0.88 | 0.81 |
| Wheat | 0.84 | 0.78 | 0.76 | 0.89 | 0.73 | 0.75 | 0.94 | 0.85 | 0.87 |
| Overall | 0.89 | | | 0.86 | | | 0.92 | | |

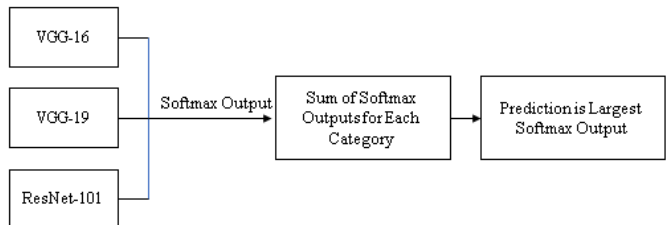

Fig. 5: Workflow of Ensemble Model

Table 2: Evaluation Results on Ensemble Model

|  | Ensemble Model | | |
| --- | --- | --- | --- |
|  | Accuracy | Precision | Recall |
| Canola | 0.94 | 0.84 | 0.90 |
| Rough Rice | 0.95 | 0.88 | 0.89 |
| Sorghum | 0.94 | 0.91 | 0.82 |
| Soy | 0.95 | 0.83 | 0.96 |
| Wheat | 0.95 | 0.88 | 0.90 |
| Overall | 0.94 | | |

## VII. CONCLUSION AND FUTURE WORK

Proposed is a seed classification framework for seed phenotyping using deep learning and synthetic image datasets. A prototypical seed phenotyping environment is used to demonstrate the feasibility of low-altitude UAV imagery and synthetic datasets to address the profusion of training data required to train neural network architectures. The use of transfer learning and fine-tuning is employed to efficiently conduct the task of classification. Besides, an ensemble model is built to improve upon the quality of predictions made by the neural networks. The next steps include leveraging a drone that is sturdy enough for stable outdoor flights and testing the framework in an outdoor environment where the acquired drone imagery is complex to work with due to varying backgrounds. Also, the impact of drone image resolution and quality on the generated synthetic image datasets is to be studied.